\documentclass[conference]{IEEEtran}
\IEEEoverridecommandlockouts

\usepackage{cite}
\usepackage{amsmath,amssymb,amsfonts}
\usepackage{algorithmic}
\usepackage{graphicx}
\usepackage{textcomp}
\usepackage{xcolor}
\usepackage[inline]{enumitem}
\usepackage{booktabs}
\usepackage{hyperref}

\def\BibTeX{{\rm B\kern-.05em{\sc i\kern-.025em b}\kern-.08em
    T\kern-.1667em\lower.7ex\hbox{E}\kern-.125emX}}

\newcommand{\authorblock}[2]{%
  \begin{minipage}[t]{0.27\textwidth}
    \centering
    \textbf{#1}\par
    #2
  \end{minipage}%
}

\begin{document}

\title{Larger Language Models Don't Care How You Think: Why Chain-of-Thought Prompting Fails in Subjective Tasks
\thanks{Funded in part by DARPA under contract HR001121C0168}
}


\author{%
\hspace{-25px}
  \authorblock{Georgios Chochlakis}{}%
  \hfill
  \authorblock{Niyantha Maruthu Pandiyan}{}%
  \hfill
  \authorblock{Kristina Lerman}{}%
  \hfill
  \authorblock{Shrikanth Narayanan}{}%
  \\
  CS and ECE - School of Advanced Computing, University of Southern California
  \\
  \textit{\{chochlak, maruthup, lerman, shri\}@usc.edu}
}

\maketitle

\begin{abstract}
In-Context Learning (ICL) in Large Language Models (LLM) has emerged as the dominant technique for performing natural language tasks, as it does not require updating the model parameters with gradient-based methods. ICL promises to ``adapt'' the LLM to perform the present task at a competitive or state-of-the-art level at a fraction of the computational cost. ICL can be augmented by incorporating the reasoning process to arrive at the final label explicitly in the prompt, a technique called Chain-of-Thought~(CoT) prompting.
However, recent work has found that ICL relies mostly on the retrieval of task priors and less so on ``learning'' to perform tasks, especially for complex subjective domains like emotion and morality, where priors ossify posterior predictions.
In this work, we examine whether ``enabling'' reasoning also creates the same behavior in LLMs, wherein the format of CoT retrieves \textit{reasoning priors} that remain relatively unchanged despite the evidence in the prompt. We find that, surprisingly, CoT indeed suffers from the same posterior collapse as ICL for larger language models. Code is avalaible at \url{https://github.com/gchochla/cot-priors}.
\end{abstract}

\section{Introduction}
\label{sec:intro}

Large Language Models (LLMs) \cite{radford2019language, ouyangTrainingLanguageModels2022, touvronLlamaOpenFoundation2023, dubey2024llama, brownLanguageModelsAre2020, achiam2023gpt} have come to dominate language processing tasks as tools that are reliable, affordable, and scalable in many disciplines~\cite{ziems2023can, bakker2022fine}. Their proliferation comes from their ``emergent'' In-Context Learning (ICL) ability \cite{brownLanguageModelsAre2020, wei2022emergent}, i.e., performing tasks by conditioning on input-output demonstrations and/or task instructions.

While ICL is often contrasted with traditional gradient-based updates of the models' parameters (also referred to as in-weights learning) \cite{kossen2023context, chanDataDistributionalProperties2022}, the ICL abilities of LLMs depend on their strong prior knowledge of the task to perform it in a zero- or few-shot manner. Therefore, studying the interplay between ICL and in-weights learning is important for our understanding of the behavior of ICL.

Prior work found evidence that LLMs may be overly reliant on their prior knowledge, disregarding the demonstrations in the prompt. Specifically, \cite{min2022rethinking} demonstrated that LLMs can ignore the provided evidence in their instructions, and instead perform ``Task Recognition''; they focus on the examples and the labels \textit{separately} to fetch the underlying task \cite{xie2021explanation}. To show this, they sampled examples and labels independently, with minuscule impact on performance. Since no ground truth is provided, the authors consider this setting as ``zero-shot'' inference, namely \textbf{task-recognition zero-shot}. While follow-up work~\cite{kossen2023context, wei2023larger, pan2023context} has further studied this phenomenon and challenged some of the original findings, more recent work~\cite{chochlakis2024strong} has provided further, quantitative evidence for the \textit{pull} these task priors exert on the posterior predictions. In particular, they find that, for \textit{complex subjective} tasks like multilabel emotion recognition, LLMs rely almost exclusively on their task priors for their posterior predictions, performing significantly worse than traditional approaches when these priors are not congruent with the ground truth of the dataset. Here, we use \textit{complex subjective} to denote tasks with multiple interrelated labels for which people can reasonably disagree about, where ``ground'' truth is substituted by crowd truth~\cite{aroyo2015truth}.  

In this work, we expand the scope to incorporate morality to the list of subjective multilabel tasks. We use experts to generate reasoning chains for the examples in the prompt and thoroughly study whether we can overcome the pull of the priors with Chain-of-Thought (CoT) prompting~\cite{weiChainThoughtPrompting2022} in six state-of-the-art LLMs. First, we evaluate how well CoT performs compared to regular ICL. Second, we design potential CoT priors and evaluate their properties. Finally, we evaluate the reasonableness of the reasoning generated by the LLMs. Surprisingly, we identify trends in CoT that are identical to those of ICL~\cite{chochlakis2024strong}, suggesting CoT is {\em not} sufficient to overcome the pull of the priors, since:
\begin{itemize}
    \item LLMs with CoT perform at the same, subpar levels with ICL in subjective tasks, such as emotions and morality,
    \item Larger and more capable LLMs indeed have \textbf{reasoning priors} that are elicited by CoT irrespective of the evidence in the prompt and ossify posterior predictions,
    \item LLM prior reasoning chains remain reasonable and coherent despite the noise introduced during inference to elicit the priors of the model.
\end{itemize}

\section{Related Work}

\subsection{ICL and the Pull of the Priors}

Since the introduction of ICL~\cite{brownLanguageModelsAre2020} as an inference technique, it has been widely used for evaluations on standard benchmarks~\cite{srivastava2022beyond}. It requires no finetuning, which is costly to perform for large models, and usually achieves competitive or state-of-the-art performance. Combined with the existence of  APIs~\cite{achiam2023gpt} and open-source implementations and weights \cite{touvronLlamaOpenFoundation2023, dubey2024llama}, ICL has become an accessible jack of all trades.

Researchers have studied many aspects of ICL, such as contrasting ICL and in-weights learning by controlling the distribution of data~\cite{chanDataDistributionalProperties2022}, examining how to best select examples for the prompt~\cite{rubin2022learning}, integrating instructions explicitly during training~\cite{touvronLlamaOpenFoundation2023}, etc. Relatedly, researchers have also tried to extract the priors of the models by providing random labels for the examples of the prompt~\cite{min2022rethinking}. By showing minimal variations in performance, these experiments suggested that LLMs recognize the task in the prompt more so than learn from it, and thereafter perform inference using their prior knowledge of the task. Since no annotations are required for such a setting, the authors suggest that this inference mode can serve as a better, less naive ``zero-shot'' baseline.
Subsequent results challenged the view that LLMs mostly perform task recognition, showcasing a significant degradation in performance when increasing the number of randomized examples in the prompt~\cite{kossen2023context}, and analyzed LLM behavior when substituting or permuting the labels~\cite{kossen2023context, pan2023context, wei2023larger}. More recent work, however, suggests that in complex subjective tasks like emotion recognition, ``Task Recognition'' dominates posterior predictions~\cite{chochlakis2024strong}. The implication is that ICL is unable to incorporate divergent perspectives.

\subsection{Chain of Thought}

One potential way to augment ICL is with CoT~\cite{weiChainThoughtPrompting2022}. CoT incorporates the derivation process explicitly in the prompt, presenting a more human-like reasoning process. This has several advantages, like making the implicit associations in the data explicit in the prompt, making responses more explainable because of the generation of the reasoning by the LLM, and directing more computing resources towards more complex problems (e.g., by producing more tokens for more complex reasoning steps). Indeed, CoT improves the robustness of prompting, and generally improves performance and the reasoning capabilities of LLMs. Subsequent work~\cite{yao2024tree} has tried to expand on CoT to enable the model to backtrack. In this work, we study whether CoT can alleviate the posterior's collapse to the prior described above~\cite{chochlakis2024strong}, as incorporating the reasoning chain explicitly could bridge the gap between priors and evidence. We also evaluate whether CoT has priors, and how \textit{reasonable} the generated reasoning is (more details in Section \ref{sec:reasonable}). Relevant to the evaluation of the reasoning of LLMs is work examining the faithfulness of CoT~\cite{lanham2023measuring, turpin2024language}.

\section{Methodology}

We closely follow the methodology and notation in \cite{chochlakis2024strong}. For a set of examples $\mathcal{X}$, and a set of labels $\mathcal{Y}$, a dataset $\mathcal{D}$ defines a mapping $f:\mathcal{X} \rightarrow \mathcal{Y}$, as well as reasoning chains $R(x) = r, x\in\mathcal{X}$, explicitly describing $f$, and therefore \mbox{$\mathcal{D} = \{(x, y, r): x \in \mathcal{X}, y = f(x), r = R(x)\}$}, from which we can sample demonstrations with $p(x, y, r)$. We do not differentiate between splits for brevity. Given CoT prompt \mbox{$S = \{(x_i, y_i, r_i): (x_i, y_i, r_i) \sim p_S, i \in [k]\}$} with $k$ demonstrations from sampling distribution $p_S$, an LLM produces its own mapping and predictions for the task, denoted as \mbox{$\hat{f}_k(.; p_S): \mathcal{X} \rightarrow \mathcal{Y}$}. For all our experiments, we set the temperature to 0 to derive deterministic predictions.

\subsection{Performance and Similarity Measures}

To evaluate both API-based and open-source LLMs, we rely on similarity measures calculated directly on the final predictions rather than probabilistic measures like the models' output logits. Using probabilistic measures is also not straightforward for multilabel tasks, as described in \cite{chochlakis2024strong}. Therefore, we use the Jaccard Score, Micro and Macro F1 metrics~\cite{mohammad2018semeval} to evaluate the performance of the models. For consistency, we also use them to quantify the similarity between the prediction sets from different model runs, since they are symmetric functions, allowing us to apply them to interchangeable sets.

\subsection{Task Prior Proxies via Task Recognition}


We use 3 CoT settings where the data for the prompt are sampled randomly. First, we have the true \textbf{reasoning task-recognition zero-shot prior}, where the prompt contains $k$ demonstrations sampled with \mbox{$p^I(x, y, r) = p(x) p(y) p(r)$}, so text, labels, and reasoning are sampled \textit{independently} from each other from $\mathcal{D}$, hence labels and reasoning are irrelevant to the text and each other. This effectively maintains the higher-order relationships between labels, which are strong in such multilabel tasks~\cite{chochlakisLeveragingLabelCorrelations2023}. We also construct two more settings for more fine-grained evaluations, where we sample only the reasoning or only the label independently with \mbox{$p^{I, r}(x, y, r) = p(x, y) p(r)$} and \mbox{$p^{I, y}(x, y, r) = p(x, r) p(y)$} respectively. We will refer to $\hat{f}_{k}(.; p^I)$, $\hat{f}_{k}(.; p^{I, r})$ and $\hat{f}_{k}(.; p^{I, y})$ as \textbf{task priors} henceforth. When using regular ICL, we simply drop the reasoning text.

\begin{figure*}
    \hspace{-15px}
    \includegraphics[scale=0.5]{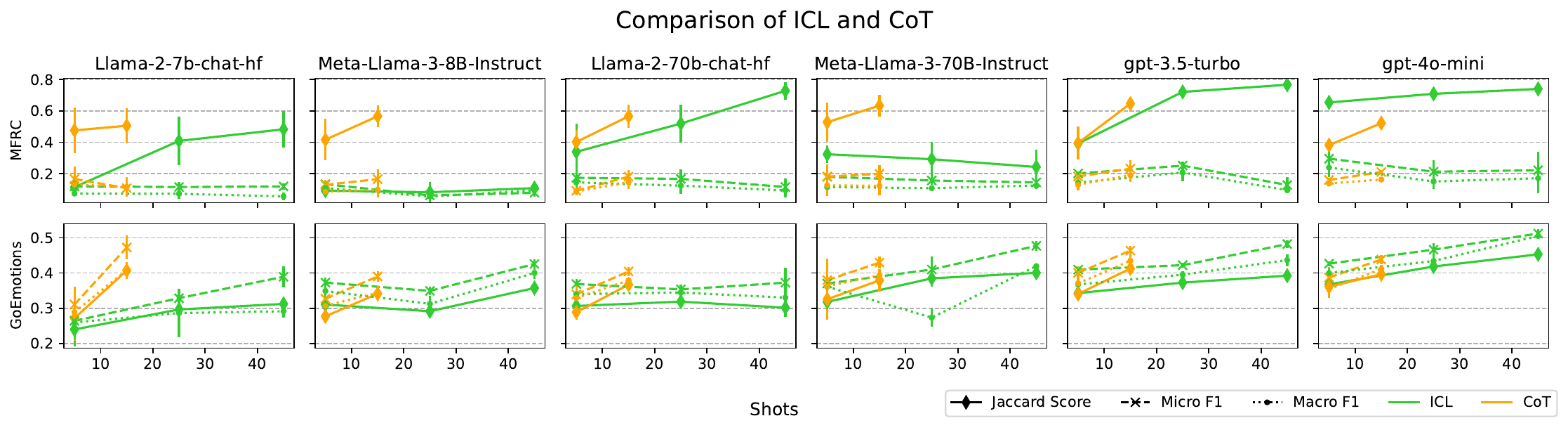}
    \caption{Performance comparison between Chain-of-Thought prompting and In-context Learning for various shots.}
    \label{fig:cot-icl}
\end{figure*}

\subsection{Pull of Prior}

To quantify the pull of the prior, we compare the similarity of $\hat{f}_k(.; p)$ with the ground truth (that is, the CoT performance of the model) and with the \textit{task priors}. Higher similarity with the task prior indicates a greater pull of priors on the final posterior predictions. To compare across models, we can use the difference between the two similarities for each.

\subsection{Reasonableness of Reasoning} \label{sec:reasonable}

We also measure the reasonableness and the plausibility of the reasoning chains produced by the LLMs, as well as how coherent and reasonable the prediction is given the generated reasoning. More concretely, we first manually evaluate whether the produced reasoning chain is relevant and describes aspects of the specific input in a plausible manner. For example, missing potential sarcasm in a document could result in erroneous rationale from the model, yet the reasoning might still be coherent and relevant to the input. Then,  irrespective of the input, we manually evaluate whether the labels can be \textit{directly} derived from the reasoning chain. We stress, therefore, that we do not evaluate the correctness of reasoning and predictions, which is rather assessed by the model's performance.

\subsection{Prompt Design}

Previous work has demonstrated that small changes in the prompt can have a significant impact on the outputs of LLMs. In our effort to reduce the search space, we standardize the prompt template, presenting the one that yields the best performance with respect to the ground truth among the ones we experimented with. In addition, because the specific examples and their order in the prompt can affect the output of the model, we use exactly the same examples, in the same order across corresponding experiments.

\section{Experiments}

\begin{figure*}
    \hspace{-15px}
    \includegraphics[scale=0.53]{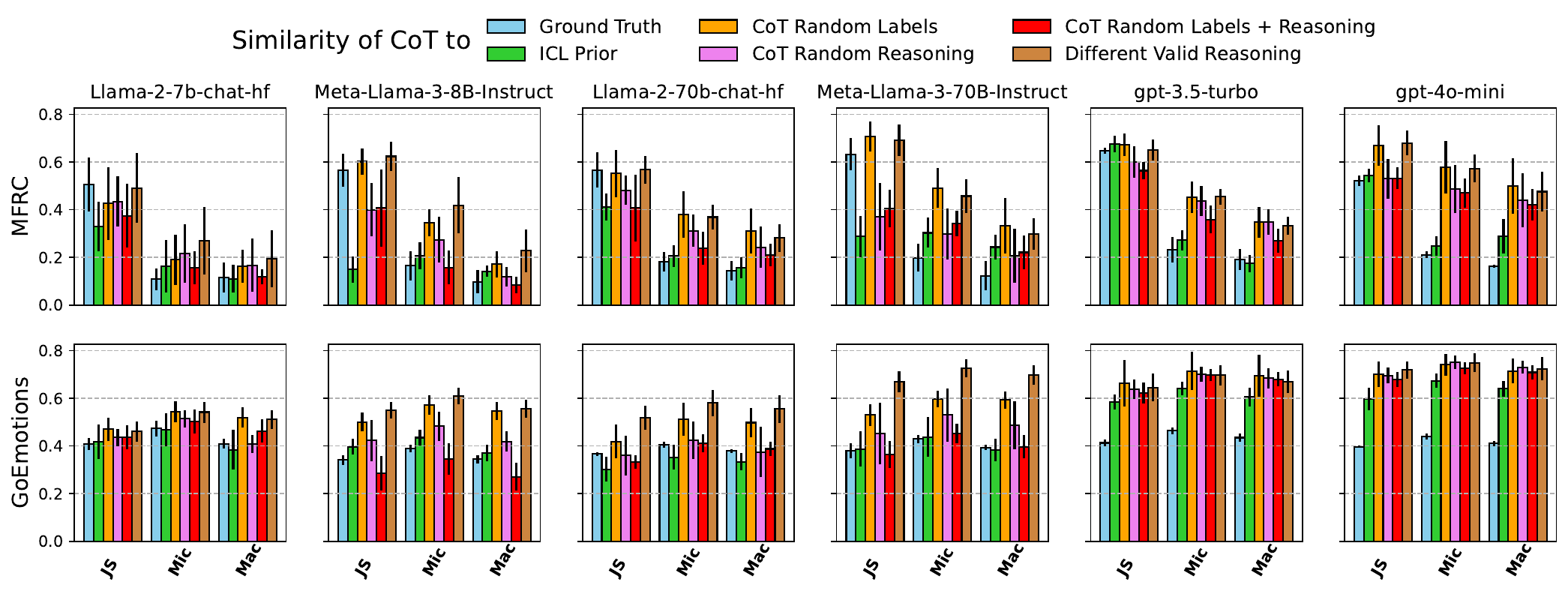}
    \caption{Similarity of Chain-of-Thought prompting to ground truth, potential task priors, and consistency across different but valid reasoning chains (\textbf{JS}: Jaccard Score, \textbf{Mic}: Micro F1, \textbf{Mac}: Macro F1).}
    \label{fig:prior}
\end{figure*}

\subsection{Datasets}

\noindent \textbf{MFRC}~\cite{trager2022moral}: Multilabel moral foundation corpus with annotations for six moral foundations: \textit{care}, \textit{equality}, \textit{proportionality}, \textit{loyalty}, \textit{authority}, and \textit{purity}. To reduce inference costs, we use a random subset of 100 examples for evaluation.

\noindent \textbf{GoEmotions} \cite{demszky2020goemotions}: Multilabel emotion recognition benchmark with 27 emotions. For consistency, we evaluate the model on a random subset of 100 examples. To make the task feasible for the LLMs, we pool the emotions to the following seven ``clusters'' by using hierarchical clustering~\cite{demszky2020goemotions}: \textit{admiration}, \textit{anger}, \textit{fear}, \textit{joy}, \textit{optimism}, \textit{sadness}, and \textit{surprise}.

\subsection{Implementation Details}

We use the 4-bit quantized versions of the open-source LLMs through the \textit{HuggingFace} interface for \textit{PyTorch}. We use LLaMA-2 (\texttt{meta-llama/Llama-2-\#b-chat-hf}), and 3 (\texttt{meta-llama/Meta-Llama-3-\#B-Instruct}), GPT-3.5 Turbo (\texttt{gpt-3.5-turbo}), and GPT-4o mini (\texttt{gpt-4o-mini}). We perform 3 runs for each LLM experiment, varying the examples used and their labels. We control which examples or labels are selected for each run to ensure consistency across models. We present mean and standard deviation. When computing them for similarities, we use every possible pair between two configurations. We use random retrieval of examples. We use less shots for CoT ($\{5, 15\}$ compared to $\{5, 25, 45\}$) given that the reasoning in the prompt increases the length of the prompt. Two experts generated reasonings independently given a single example, and annotated the reasonableness of the generated reasoning chains by the LLMs. We use the reasonings of one annotator for our experiments, and use the other's to check consistency.

\subsection{Performance gains from CoT}

First, we evaluate whether CoT can improve the performance of LLMs above and beyond ICL. In Fig. \ref{fig:cot-icl}, we present the performance of ICL, and compare it to CoT.

First, for MFRC, we note that performance w.r.t. Jaccard Score is much higher compared to the F1 scores for both methods. Given the label sparsity, this disparity in values indicates that the model is struggling with true positive, making the Jaccard Score suboptimal for comparison between models. Therefore, we focus our analysis on the F1 scores. It is evident that CoT does not present any improvement relative to ICL, which itself is a weak baseline. We see that performance does not significantly differ across models. An exception is GPT-4o, where 5-shot ICL achieves notably better scores, yet performance degrades with more examples. In fact, it does so consistently with more ICL examples.

For GoEmotions, on the other hand, we notice that all metrics have similar values, and therefore can focus our analysis on any to derive insights. First, we note that we indeed see scaling of the performance across models, with smaller and/or capable models performing the worst, as expected. Nonetheless, when integrating CoT, we see that the best performing models are not really augmented, whereas the smallest model improves radically. This greater malleability of smaller models is consistent with the findings in \cite{chochlakis2024strong}. 

Overall, we find benefits from CoT only in smaller models, and very little improvement otherwise. Therefore, we conclude that CoT does not improve performance in complex subjective tasks, especially for the latest models.

\subsection{Reasoning Priors}

To analyze the reasons behind CoT failing to improve performance, we first look at whether the models have \textbf{reasoning priors} that ossify posterior predictions despite the evidence presented to the model in the form of proper reasoning chains and labels. In Fig. \ref{fig:prior}, we present the similarity of the predictions of CoT ($\hat{f}_{15}(.; p)$) with
\begin{enumerate*}[label={(\roman*)}]
    \item the ground truth $f$ (and therefore the performance of the modes),
    \item the ICL prior $\hat{f}_{25}(.; p^I)$,
    \item CoT with random labels $\hat{f}_{15}(.; p^{I, y})$,
    \item CoT with random reasoning chains $\hat{f}_{15}(.; p^{I, r})$,
    \item CoT with both random reasoning chains and labels $\hat{f}_{15}(.; p^{I})$, which would be the proper prior for CoT, and
    \item CoT using the reasoning chains of another annotator to check for consistency.
\end{enumerate*}
 
Considering the potential of CoT, our results are surprising yet consistent with previous findings on ICL~\cite{chochlakis2024strong}. We notice that in less capable models, randomizing the reasoning chains (with or without random labels) decreases the similarity to CoT, which usually remains lower than the similarity to the ground truth and the consistency of CoT. It is interesting to see that randomizing only the labels does not significantly impact predictions, as similarity tends to be on par with consistency. However, for bigger and more capable models, like \mbox{LLaMA-3} and GPT, it becomes evident that \textit{LLMs develop reasoning priors}. In particular, the similarity to the (potential) CoT prior becomes notably larger than that to the ground truth, and even the ICL prior starts to resemble CoT more than the ground truth. For reference, we note that the similarities of the ICL prior to the CoT prior and the regular COT are similar in value.

\subsection{Coherence of Reasoning Chains}

\begin{table}[]
    \centering
    \caption{Reasonableness of reasoning $r$, and that of label $y$ given $r$, where $r$ and $y$ are generated in response to query $x$.}
    \begin{tabular}{llcc}
        && \multicolumn{2}{c}{\textbf{Reasonable?}} \\
        \cmidrule{3-4}
        \textbf{Model} & \textbf{Dataset} & Reasoning & Label \\
        \midrule
        GPT-4o & MFRC & 99.0\% & 87.0\% \\
        GPT-4o & GoEmotions & 95.0\% & 92.0\% \\
        Llama 3 70B & MFRC & 92.0\% & 91.0\% \\
        Llama 3 70B & GoEmotions & 90.0\% & 98.0\% \\
        GPT-4o prior & MFRC & 81.0\% & 48.0\% \\
        GPT-4o prior & GoEmotions & 89.0\% & 94.0\% \\
        
    \end{tabular}
    \label{tab:reasonableness}
\end{table}

For a complete analysis of CoT, but also to further demonstrate the presence of \textit{reasoning priors}, we manually evaluate the reasonableness of the reasoning chains and the labels generated both by CoT and the CoT prior. We present our findings in Table \ref{tab:reasonableness} for GPT-4o and its prior, and LLaMA-3 for both datasets.
It is interesting to see that for both CoT and its prior, the level of reasonableness is quite high in both reasoning and labels, except for MFRC label coherence. Since the level of reasonableness is quite high for the randomized prompts, this indicates that the randomization of reasoning and labels truly evokes the reasoning prior of the model. Finally, from our manual evaluation, we notice that the LLMs routinely miss nuanced meanings, such as sarcasm, yet their more rudimentary reasoning still remains valid.

\section{Conclusion}

In this work, we evaluate whether CoT displays the same priors as simple ICL. We find, surprisingly, that larger language models do have \textbf{reasoning priors} that ossify their generated reasoning and posterior predictions, virtually disregarding the evidence in the form of reasoning and labels provided in the prompt.
Given that the performance of LLMs is inferior to traditional methods in the tasks we study, this is unlikely to be an artifact of data memorization.
We conclude that reasoning in the form of CoT is unable to overcome the pull of the priors in larger language models, leading to suboptimal performance on subjective tasks in the form of multilabel benchmarks, as opposed to other tasks where labels, and in turn reasoning, remain more consistent across datasets, the validity of a label can be asserted more formally and consistently, and therefore agreement is much higher.

\vfill
\pagebreak

\bibliographystyle{IEEEtran}
\bibliography{references}

\end{document}